\documentclass[11pt,letterpaper]{article}
\usepackage{naaclhlt2016}
\usepackage{times}
\usepackage{latexsym}

\usepackage{amsfonts}
\usepackage{amsmath}
\usepackage{amssymb}
\usepackage{longtable}
\usepackage{graphicx}

\usepackage{tabularx}
\usepackage{arydshln}
\usepackage{lipsum}
\usepackage{booktabs}

\usepackage{epstopdf}
\DeclareGraphicsExtensions{.pdf,.eps,.png} 
\usepackage{multirow}

\newcommand{\matr}[1]{\mathbf{#1}} 

\newcolumntype{L}{>{\centering\arraybackslash}m{1.5cm}}

\naaclfinalcopy 
\def\naaclpaperid{455} 

\title{Brundlefly at SemEval-2016 Task 12:  \\ Recurrent Neural Networks vs. Joint Inference \\ for Clinical Temporal Information Extraction }

\author{Jason Alan Fries\\
	    Stanford University\\
	    Mobilize Center\\
	    {\tt jason-fries@stanford.edu}
}
\date{}

\begin{document}

\maketitle

\begin{abstract}
We submitted two systems to the SemEval-2016 Task 12: Clinical TempEval challenge, participating in Phase 1, where we identified text spans of time and event expressions in clinical notes and Phase 2, where we predicted a relation between an event and its parent document creation time. 

For temporal entity extraction, we find that a joint inference-based approach using structured prediction outperforms a vanilla recurrent neural network that incorporates word embeddings trained on a variety of large clinical document sets. For document creation time relations, we find that a combination of date canonicalization and distant supervision rules for predicting relations on both events and time expressions improves classification, though gains are limited, likely due to the small scale of training data. 
\end{abstract}

\section{Introduction}

This work discusses two information extraction systems for identifying temporal information in clinical text, submitted to SemEval-2016 Task 12 : Clinical TempEval  \cite{bethard-EtAl:2016:SemEval}. We participated in tasks from both phases: (1) identifying text spans of time and event mentions; and (2) predicting relations between clinical events and document creation time. 

Temporal information extraction is the task of constructing a timeline or ordering of all events in a given document. In the clinical domain, this is a key requirement for medical reasoning systems as well as longitudinal research into the progression of disease. While timestamps and the structured nature of the electronic medical record (EMR) directly capture some aspects of time, a large amount of information on the progression of disease is found in the unstructured text component of the EMR where temporal structure is less obvious.

We examine a deep-learning approach to sequence labeling using a vanilla \emph{recurrent neural network} (RNN) with word embeddings, as well as a joint inference, structured prediction approach using Stanford's knowledge base construction framework DeepDive \cite{zhang2015deepdive}.
Our DeepDive application outperformed the RNN and scored similarly to 2015's best-in-class extraction systems, even though it only used a small set of context window and dictionary features. Extraction performance, however lagged this year's best system submission. For document creation time relations, we again use DeepDive. Our system examined a simple temporal distant supervision rule for labeling time expressions and linking them to nearby event mentions via inference rules.  Overall system performance was better than this year's median submission, but again fell short of the best system.

\section{Methods and Materials}
\label{sec:methods}


Phase 1 of the challenge required parsing clinical documents to identify  {\sc Timex3} and {\sc Event} temporal entity mentions in text. {\sc Timex3} entities are expressions of time, ranging from concrete dates to phrases describing intervals like ``the last few months."  {\sc Event} entities are broadly defined as anything relevant to a patient's clinical timeline, e.g., diagnoses, illnesses, procedures.  Entity mentions are tagged using a document collection of  clinic and pathology notes from the Mayo Clinic called the THYME (Temporal History of Your Medical Events) corpus \cite{styler2014temporal}.

We treat Phase 1 as a sequence labeling task and examine several models for labeling entities. We discuss our submitted tagger which uses a vanilla RNN\footnote{There was a bug in our original RNN submission that dramatically lowered recall. We report the original and corrected test set scores in order to better characterize the contributions of this work.}  and compare its performance to a DeepDive-based system, which lets us encode domain knowledge and sequence structure into a probabilistic graphic model.

For Phase 2, we are given all test set entities and asked to identify the temporal relationship between an {\sc Event} mention and corresponding document creation time. This relation is represented as a classification problem, assigning event attributes from the label set  {\sc \{Before, Overlap, Before/Overlap, After\}}. We use DeepDive to define several inference rules for leveraging neighboring pairs of {\sc Event} and {\sc Timex3} mentions to better reason about temporal labels. 

\subsection{Recurrent Neural Networks}

\subsubsection{Overview}
\begin{figure}[!htb]
\center
\includegraphics[width=75mm]{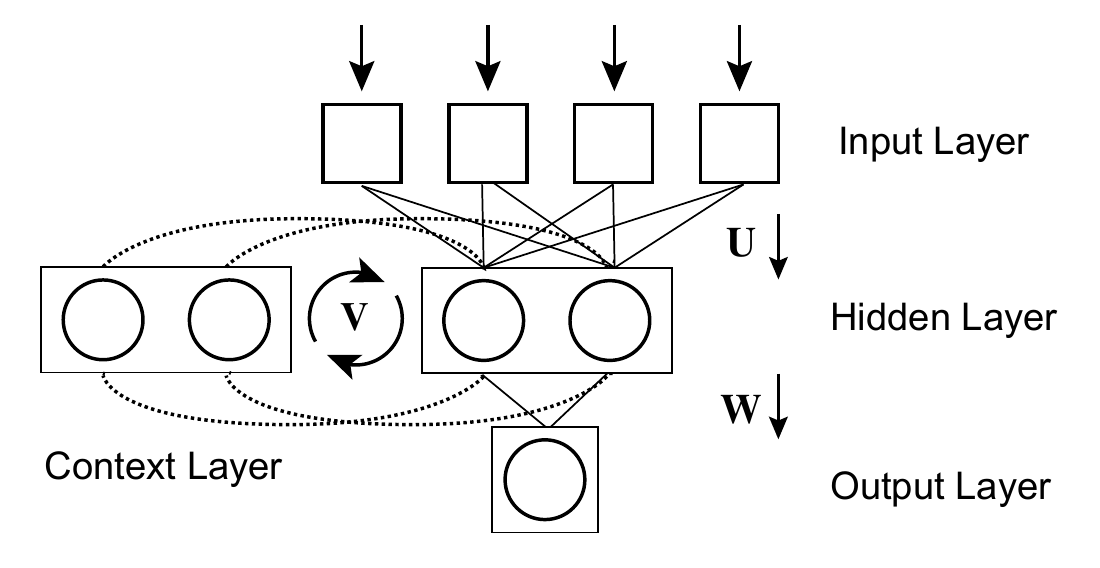}
\footnotesize
\caption{Simple Recurrent Neural Network. $\matr{U}$ is the input $\times$ hidden layer weight matrix. $\matr{V}$ is the context layer $\times$ hidden layer matrix, and $\matr{W}$ is the output weight matrix.  Dotted lines indicate recurrent edge weights.}
\label{fig:rnn-diagram}
\end{figure}

Vanilla (or Elman-type) RNNs are recursive neural networks with a linear chain structure \cite{elman1990finding}. RNNs are similar to classical feedforward neural networks, except that they incorporate an additional hidden context layer that forms a time-lagged, recurrent connection (a directed cycle) to the primary hidden layer. In the canonical RNN design, the output of the hidden layer at time step $t-1$ is retained in the context layer and fed back into the hidden layer at $t;$ this enables the RNN to explicitly model some aspects of sequence history. (see Figure \ref{fig:rnn-diagram}). 

Each word in our vocabulary is represented as an $n$-dimensional vector in a lookup table of $|vocab|$ x $n$ parameters (i.e., our learned embedding matrix). Input features then consist of a concatenation of these embeddings to represent a context window surrounding our target word. The output layer then emits a probability distribution in the dimension of the candidate label set. The lookup table is shared across all input instances and updated during training.  

Formally our RNN definition follows \cite{mesnil2013investigation}:
\begin{center}
$h(t) = f(\matr{U}x(t) + \matr{V}h(t-1))$
\end{center}
where $x(t)$ is our concatenated context window of word embeddings, $h(t)$ is our hidden layer, $\matr{U}$ is the input-to-hidden layer matrix, $\matr{V}$ is the hidden layer-to-context layer matrix, and $f(x)$ is the activation function (logistic in this work).
\begin{center}
$f(x) =  \frac{1}{1+e^{-x}}$
\end{center}
The output layer $y(t)$ consists of a softmax activation function $g(x)$
\begin{center}
$y(t) = g(\matr{W}h(t))$ \ \ \ \ \ $g(x_i) =  \frac{e^{x_i}}{  \sum_{k}  e^{x_{k}}  }$
\end{center}
where $\matr{W}$ is the output layer matrix. Training is done using batch gradient descent using one sentence per batch. Our RNN implementation is based on code available as part of Theano v0.7 \cite{Bastien-Theano-2012}.

\subsubsection{Word Embeddings}

For baseline RNN models, all embedding parameters are initialized randomly in the range [-1.0, 1.0]. For all other word-based models, embedding vectors are initialized or \emph{pre-trained} with parameters trained on different clinical corpora. Pre-training generally improves classification performance over random initialization and provides a mechanism to leverage large collections of unlabeled data for use in semi-supervised learning \cite{erhan2010does}. 

\begin{table}[!ht]
\small
\centering
\begin{tabular}{|l|l|c|c|c|c|c|} \hline
\bf Corpus	& \bf Note Type	&	\bf Tokens	& \bf $|$Vocab$|$ \\ \hline
MIMIC-III	&	Mixed (15 types)	&	758M	&	260K \\
UIHC-ALL	 & Mixed (41 types) &	6.5B	&	899K \\
UIHC-CN	&	Clinic Notes	& 2.3B	&	378K \\
UIHC-PN	&	Progress Notes	&	1.8B	&	378K \\
\hline
\end{tabular}
\caption{\label{tbl:emb-corpora} Summary statistics for embedding corpora.}
\end{table}

We create word embeddings using two collections of clinical documents: the MIMIC-III database containing 2.4M notes from critical care patients at Beth Israel Deaconess Medical Center \cite{saeed2011multiparameter}; and the University of Iowa Hospitals and Clinics (UIHC) corpus, containing 15M predominantly inpatient notes (see Table \ref{tbl:emb-corpora}).
All word embeddings in this document are trained with word2vec \cite{mikolov2013efficient} using the Skip-gram model, trained with a 10 token window size. We generated 100 and 300 dimensional embeddings based on prior work tuning representation sizes in clinical domains \cite{fries2015dissertation}. 

\subsubsection{RNN Taggers}
We train RNN models for three tasks in Phase 1: a character-level RNN for tokenization; and two word-level RNNs for POS tagging and entity labeling. Word-level RNNs are pre-trained with the embeddings described above, while character-level RNNs are randomly initialized. All words are normalized by lowercasing tokens and replacing digits with {\tt N}, e.g., {\tt 01-Apr-2016} becomes {\tt NN-apr-NNNN} to improve generalizability and restrict vocabulary size. Characters are left as unnormalized input. In the test data set, unknown words/characters are represented using the special token {\tt<UNK>} .  All hyperparameters were selected using a randomized grid search. \footnote{RNN parameter space: hidden units: [48 - 384]; L/R context window [2 - 6]; learning rate: (0.001, 0.01, 0.1, 0.25). Character-level RNN parameters: dim: (16, 32); sentence padding [0 - 5]. }

{\bf Tokenization}: Word tokenization and sentence boundary detection are done simultaneously using a character-level RNN. Each character is assigned a tag from 3 classes: {\tt WORD}{\tt(W)} if a character is a member of a token that does not end a sentence; {\tt END}{\tt(E)} for a token that does end a sentence, and whitespace {\tt O}. We use IOB2 tagging to encode the range of token spans. 

Models are trained using THYME syntactic annotations from colon and brain cancer notes. Training data consists of all sentences, padded with 5 characters from the left and right neighboring sentences. Each character is represented by a 16-dimensional embedding (from an alphabet of 90 characters) and an 11 character context window. The final prediction task input is one, long character sequence per-document. We found that the tokenizer consistently made errors conflating E and W classes (e.g., {\tt B-W, I-E, I-E}) so after tagging, we enforce an additional consistency constraint on {\tt B-*} and  {\tt I-*} tags so that contiguous BEGIN/INSIDE spans share the same class.  

{\bf Part-of-speech Tagging}: We trained a POS tagger using THYME syntactic annotations. A model using 100-dimensional UIHC-CN embeddings (clinic notes) and a context window of $\pm$ 2 words performed best on held out test data, with an accuracy of 97.67\% and F$_1$= 0.973.

{\bf TIMEX3 and EVENT  Span Tagging}:  We train separate models for each entity type, testing different pre-training schemes using 100 and 300-dimensional embeddings trained on our large, unlabeled clinical corpora. Both tasks use context windows of $\pm$ 2 words (i.e., concatenated input of 5 $n$-d word embeddings) and a learning rate of 0.01. We use 80 hidden units for 100-dimensional embeddings models and 256 units for 300-dimensional models. Output tags are in the IOB2 tagging format.

\subsection{DeepDive}

DeepDive developers build domain knowledge into applications using a combination of \emph{distant supervision rules}, which use heuristics to generate noisy training examples, and inference rules which use \emph{factors} to define relationships between random variables. This design pattern allows us to quickly encode domain knowledge into a probabilistic graphical model and do joint inference over a large space of random variables. 

For example, we want to capture the relationship between {\sc Event} entities and their closest {\sc Timex3} mentions in text since that provides some information about when the {\sc Event} occurred relative to document creation time. {\sc Timex3}s lack a class {\tt DocRelTime}, but we can use a distant supervision rule to generate a noisy label that we then leverage to predict neighboring  {\sc Event} labels. We also know that the set of all {\sc Event}/{\sc Timex3} mentions within a given note section, such as patient history, provides discriminative information that should be shared across labels in that section. DeepDive lets us easily define these structures by linking random variables (in this case all entity class labels) with factors, directly encoding domain knowledge into our learning algorithm.
 
 \subsubsection{TIMEX3 and EVENT Spans}
{\bf Phase 1}:  Our baseline tagger consists of three inference rules: logistic regression, conditional random fields (CRF), and skip-chain CRF \cite{sutton2006introduction}.  In CRFs, factor edges link adjoining words in a linear chain structure, capturing label dependencies between neighboring words. Skip-chain CRFs generalize this idea to include skip edges, which can connect non-neighboring words. For example, we can link labels for all identical words in a given window of sentences. We use DeepDive's feature library, ddlib, to generate common textual features like context windows and dictionary membership. We explored combinations of left/right windows of 2 neighboring words and POS tags, letter case, and entity dictionaries for all vocabulary identified by the challenge's baseline memorization rule, i.e., all phrases that are labeled as true entities $\ge$ 50\% of the time in the training set.

{\bf Feature Ablation Tests} We evaluate 3 feature set combinations to determine how each contributes predictive power to our system. 

{\bf Run 1}: dictionary features, letter case 

{\bf Run 2}: dictionary features, letter case, context window ($\pm$ 2 normalized words)

{\bf Run 3}: dictionary features, letter case, context window ($\pm$ 2 normalized words), POS tags

\subsubsection{Document Creation Time Relations}

{\bf Phase 2}:  In order to predict the relationship between an event and the creation time of its parent document, we assign a {\tt DocRelTime} random variable to every {\sc Timex3} and {\sc Event} mention. For {\sc Event}s, these values are provided by the training data, for {\sc Timex3}s we have to compute  class labels.  Around 42\% of  {\sc Timex3} mentions are simple dates (``12/29/08", ``October 16", etc.) and can be naively canonicalized to a universal timestamp. This is done using regular expressions to identify common date patterns and heuristics to deal with incomplete dates. The missing year in ``October 16", for example, can be filled in using the nearest preceding date mention; if that isn't available we use the document creation year. These mentions are then assigned a class using the parent document's {\tt DocTime} value and any revision timestamps. Other {\sc Timex3} mentions are more ambiguous so we use a distant supervision approach. Phrases like ``currently" and ``today's"  tend to occur near {\sc Event}s that overlap the current document creation time, while ``ago" or ``$X$-years" indicate past events.  These dominant temporal associations can be learned from training data and then used to label {\sc Timex3}s.  Finally, we define a logistic regression rule to predict entity {\tt DocRelTime} values as well as specify a linear skip-chain factor over {\sc Event} mentions and their nearest {\sc Timex3} neighbor, encoding the baseline system heuristic directly as an inference rule.  

\section{Results}
\label{sec:results}

\subsection{Phase 1}

Word tokenization performance was high, F$_1$=0.993 while sentence boundary detection was lower with F$_1$ = 0.938 (document micro average F$_1$ = 0.985). Tokenization errors were largely confined to splitting numbers and hyphenated words (``ex-smoker" vs. ``ex - smoker") which has minimal impact on upstream sequence labeling. Sentence boundary errors were largely missed terminal words, creating longer sentences, which is preferable to short, less informative sequences in terms of impact on RNN mini-batches.

\begin{table}[!ht]
\small
\centering
\begin{tabular}{|l|ccc|} \hline
 \multirow{2}*{\bf Method} &  \multicolumn{3}{c|}{\bf Span}  \\ 	
  & \bf P  & \bf R  & \bf F$_1$	\\ \hline	
Baseline: memorize	&	0.774	&	0.428	&	0.551 \\ 
BluLab (2015) $^{1}$ 	&	0.797	& 0.664 &	0.725 \\ 
\textit{Median System (2016)}  	&	\textit{0.779}	& \textit{0.539} &	\textit{0.637} \\ 
\textbf{\textit{Best (2016)}}:  	& \textbf{\textit{0.840}} & \textbf{\textit{0.758}} &	\textbf{\textit{0.795}}\\ 
\hline
	
DeepDive:	 run 1 &	0.655	&	0.566	&	0.607 \\ 
DeepDive: run 2 &	0.795	&	0.675	&	\bf 0.730 \\ 
DeepDive:	 run 3	&	\bf  0.798	&	0.665	&	0.725 \\ \hline

RNN+randinit-100	&	0.694 & 0.679 & 0.686 \\ 

RNN+MIMIC-III-100	&	0.706 & 0.695 & 0.701  \\ 
RNN+UIHC-CN-100	&	0.699 & 0.688 & 0.693  \\ 
RNN+UIHC-PN-100	&	0.701 &   0.670 & 0.685  \\ 
RNN+UIHC-ALL-100	&	0.708 & 0.688 & 0.698 \\ \hline

RNN+randinit-300	&  0.713 	& 	0.657 & 	0.684 \\ 
RNN+UIHC-CN-300	&	0.704 & 0.702 & 0.703 \\ 
RNN+UIHC-PN-300	&	0.700 &  0.697 & 0.698 \\ \hline
RNN Best Ensemble	&	0.708	&	\bf 0.704	&	0.706 \\ \hline
\emph{Phase 1 RNN Submission} & 0.686 & 0.415 & 0.517 \\ 
\hline
\end{tabular}
\caption{\label{tbl:timex-table} {\sc Timex3} spans extraction performance for the test set (mean of 5 runs) [1] Baseline and BluLab scores are provided in \protect\cite{bethard2015semeval} } 
\end{table}

Tables \ref{tbl:timex-table} and \ref{tbl:event-table} contain results for all sequence labeling models. For {\sc Timex3} spans, the best RNN ensemble model performed poorly compared to the winning system (0.706 vs. 0.795). DeepDive runs 2-3 performed as well as 2015's best system, but also fell short of the top system (0.730 vs. 0.795).
{\sc Event} spans were easier to tag and RNN models compared favorably with DeepDive, the former scoring higher recall and the latter higher precision. Both approaches scored below this year's best system (0.885 vs. 0.903).

\begin{table}[!ht]
\small
\centering
\begin{tabular}{|l|ccc|} \hline
 \multirow{2}*{\bf Method} &  \multicolumn{3}{c|}{\bf Span}  \\ 	
  & \bf P  & \bf R  & \bf F$_1$	\\ \hline
Baseline: memorize	&	0.878	&	0.834	& 0.855 \\ 
BluLab (2015)  &	0.887	& 0.864 &	0.875 \\ 
\textit{Median System (2016)}  	&	\textit{0.887}	& \textit{0.846} &	\textit{0.874} \\ 
\textbf{\textit{Best (2016)}}:  	& \textbf{\textit{0.915}}	& \textbf{\textit{0.891}} & \textbf{\textit{0.903}} \\ 
\hline
DeepDive:	 run 1 &	0.864	&	0.836	&	 0.850 \\ 
DeepDive: run 2 &	\bf  0.900	&	0.864	&	0.882 \\ 
DeepDive:	 run 3	&	\bf  0.900	&	0.866	&	0.883 \\ 
\hline
RNN+randinit-100		&	0.864	&	0.869	&	0.866 \\ 
RNN+MIMIC-III-100	&	 0.861	&	0.877	&	0.869 \\ 
RNN+UIHC-CN-100	&	0.873	&	0.887	&	0.880  \\ 
RNN+UIHC-PN-100	&	0.878	&	0.882	&	0.880 \\ 
RNN+UIHC-ALL-100	&	0.878	&	0.880	&	0.879 \\ \hline
RNN+randinit-300		&	0.862	&	0.859	&	0.861 \\ 
RNN+UIHC-CN-300	&	0.880	&	0.879	&	0.879 \\ 
RNN+UIHC-PN-300	&	0.864	&	\bf 0.892	&	0.878 \\ \hline
RNN Best Ensemble	&	0.882	&	0.889	&	\bf 0.885 \\ \hline
\emph{Phase 1 RNN Submission} & 0.883 &	0.660 & 0.755 \\
\hline
\end{tabular}
\caption{\label{tbl:event-table} EVENT spans extraction performance. } 
\end{table}

\subsection{Phase 2}

Finally, Table \ref{docreltime-table} contains our DocRelTime relation extraction. Our simple distant supervision rule leads to better performance than then median system submission, but also falls substantially short of current state of the art.

\begin{table}[!ht]
\small
\centering
\begin{tabular}{|l|l|c|c|c|c|c|} \hline
\bf Attribute	& \bf P	&	\bf R & \bf F$_1$ \\ \hline
Baseline: memorize / closest	&	-	&	-	&	0.675 \\ 
BluLab (2015) 	&	-	&	-	&	0.791 \\ 
\textit{Median System (2016)} 	&	-	&	-	&	0.724 \\ 
\textbf{\textit{Best (2016)}}	&	-	&	-	&	\textbf{\textit{0.843}} \\ 
\hline
Logistic Regression (LR)	&	0.738	&	0.737	&	0.737 \\ 
LR+Skip-chain	&	0.743	&	0.742	&	\bf{0.743} \\ \hline
\end{tabular}
\caption{\label{docreltime-table} Phase 2: EVENT Document Creation Time Relation extraction measures (baseline precision/recall scores not provided). }
\end{table}

\section{Discussion}

Randomly initialized RNNs generally weren't competitive to our best performing structured prediction models (DeepDive runs 2-3)  which isn't surprising considering the small amount of training data available compared to typical deep-learning contexts. There was a statistically significant improvement for RNNs pre-trained with clinical text word2vec embeddings, reflecting the consensus that embeddings capture some syntactic and semantic information that must otherwise be manually encoded as features. Performance was virtually the same across all embedding types, independent of corpus size, note type, etc. While embeddings trained on more data perform better in semantic tasks like synonym detection, its unclear if that representational strength is important here. Similar performance might also just reflect the general ubiquity with which temporal vocabulary occurs in all clinical note contexts. Alternatively, vanilla RNNs rarely achieve state-of-the-art performance in sequence labeling tasks due to well-known issues surrounding the vanishing or exploding gradient effect \cite{pascanu2012difficulty}. More sophisticated recurrent architectures with gated units such as Long Short-Term Memory (LSTM), \cite{hochreiter1997long} and gated recurrent unit \cite{journalscorrChungGCB14} or recursive structures like Tree-LSTM \cite{tai2015improved} have shown strong representational power in other sequence labeling tasks. Such approaches might perform better in this setting.

DeepDive's feature generator libraries let us easily create a large space of binary features and then let regularization address overfitting. In our extraction system, just using a context window of $\pm$ 2 words and dictionaries representing the baseline memorization rules was enough to achieve median system performance. POS tag features had no statistically significant impact on performance in either {\sc Event}/{\sc Timex3} extraction. 

For classifying an {\sc Event}'s document creation time relation, our DeepDive application essentially implements the joint inference version of the baseline memorization rule, leveraging entity proximity to increase predictive performance. A simple distant supervision rule that canonicalizes {\sc Timex3} timestamps and predicts nearby {\sc Event}'s lead to a slight performance boost, suggesting that using a larger collection of unlabeled note data could lead to further increases. 

While our systems did not achieve current state-of-the-art performance, DeepDive matched last year's top submission for {\sc Timex3} and {\sc Event} tagging with very little upfront engineering -- around a week of dedicated development time. One of the primary goals of this work was to avoid an over-engineered extraction pipeline, instead relying on feature generation libraries or deep learning approaches to model underlying structure. Both systems explored in this work were successful to some extent, though future work remains in order to close the performance gap between these approaches and current state-of-the-art systems.

\section*{Acknowledgments}
This work was supported by the Mobilize Center, a National Institutes of Health Big Data to Knowledge (BD2K) Center of Excellence supported through Grant U54EB020405.

\bibliography{naaclhlt2016}

\begin{thebibliography}{}

\bibitem[\protect\citename{Bastien \bgroup et al.\egroup
  }2012]{Bastien-Theano-2012}
Fr{\'{e}}d{\'{e}}ric Bastien, Pascal Lamblin, Razvan Pascanu, James Bergstra,
  Ian~J. Goodfellow, Arnaud Bergeron, Nicolas Bouchard, and Yoshua Bengio.
\newblock 2012.
\newblock Theano: new features and speed improvements.
\newblock Deep Learning and Unsupervised Feature Learning NIPS 2012 Workshop.

\bibitem[\protect\citename{Bethard \bgroup et al.\egroup
  }2015]{bethard2015semeval}
Steven Bethard, Leon Derczynski, Guergana Savova, Guergana Savova, James
  Pustejovsky, and Marc Verhagen.
\newblock 2015.
\newblock Semeval-2015 task 6: Clinical tempeval.
\newblock {\em Proc. SemEval}.

\bibitem[\protect\citename{Bethard \bgroup et al.\egroup
  }2016]{bethard-EtAl:2016:SemEval}
Steven Bethard, Guergana Savova, Wei-Te Chen, Leon Derczynski, James
  Pustejovsky, and Marc Verhagen.
\newblock 2016.
\newblock Semeval-2016 task 12: Clinical tempeval.
\newblock In {\em Proceedings of the 10th International Workshop on Semantic
  Evaluation (SemEval 2016)}, San Diego, California, June. Association for
  Computational Linguistics.

\bibitem[\protect\citename{Chung \bgroup et al.\egroup
  }2014]{journalscorrChungGCB14}
Junyoung Chung, {\c{C}}aglar G{\"{u}}l{\c{c}}ehre, KyungHyun Cho, and Yoshua
  Bengio.
\newblock 2014.
\newblock Empirical evaluation of gated recurrent neural networks on sequence
  modeling.
\newblock {\em CoRR}, abs/1412.3555.

\bibitem[\protect\citename{Elman}1990]{elman1990finding}
Jeffrey~L Elman.
\newblock 1990.
\newblock Finding structure in time.
\newblock {\em Cognitive science}, 14(2):179--211.

\bibitem[\protect\citename{Erhan \bgroup et al.\egroup }2010]{erhan2010does}
Dumitru Erhan, Yoshua Bengio, Aaron Courville, Pierre-Antoine Manzagol, Pascal
  Vincent, and Samy Bengio.
\newblock 2010.
\newblock Why does unsupervised pre-training help deep learning?
\newblock {\em The Journal of Machine Learning Research}, 11:625--660.

\bibitem[\protect\citename{Fries}2015]{fries2015dissertation}
Jason Fries.
\newblock 2015.
\newblock {\em Modeling Words for Online Sexual Behavior Surveillance and
  Clinical Text Information Extraction}.
\newblock {Ph.D.} thesis, University of Iowa.

\bibitem[\protect\citename{Hochreiter and Schmidhuber}1997]{hochreiter1997long}
Sepp Hochreiter and J{\"u}rgen Schmidhuber.
\newblock 1997.
\newblock Long short-term memory.
\newblock {\em Neural computation}, 9(8):1735--1780.

\bibitem[\protect\citename{Mesnil \bgroup et al.\egroup
  }2013]{mesnil2013investigation}
Gr{\'e}goire Mesnil, Xiaodong He, Li~Deng, and Yoshua Bengio.
\newblock 2013.
\newblock Investigation of recurrent-neural-network architectures and learning
  methods for spoken language understanding.
\newblock In {\em INTERSPEECH}, pages 3771--3775.

\bibitem[\protect\citename{Mikolov \bgroup et al.\egroup
  }2013]{mikolov2013efficient}
Tomas Mikolov, Kai Chen, Greg Corrado, and Jeffrey Dean.
\newblock 2013.
\newblock Efficient estimation of word representations in vector space.
\newblock {\em arXiv preprint arXiv:1301.3781}.

\bibitem[\protect\citename{Pascanu \bgroup et al.\egroup
  }2012]{pascanu2012difficulty}
Razvan Pascanu, Tomas Mikolov, and Yoshua Bengio.
\newblock 2012.
\newblock On the difficulty of training recurrent neural networks.
\newblock {\em arXiv preprint arXiv:1211.5063}.

\bibitem[\protect\citename{Saeed \bgroup et al.\egroup
  }2011]{saeed2011multiparameter}
Mohammed Saeed, Mauricio Villarroel, Andrew~T Reisner, Gari Clifford, Li-Wei
  Lehman, George Moody, Thomas Heldt, Tin~H Kyaw, Benjamin Moody, and Roger~G
  Mark.
\newblock 2011.
\newblock Multiparameter intelligent monitoring in intensive care ii
  (mimic-ii): a public-access intensive care unit database.
\newblock {\em Critical care medicine}, 39(5):952.

\bibitem[\protect\citename{Styler~IV \bgroup et al.\egroup
  }2014]{styler2014temporal}
William~F Styler~IV, Steven Bethard, Sean Finan, Martha Palmer, Sameer Pradhan,
  Piet~C de~Groen, Brad Erickson, Timothy Miller, Chen Lin, Guergana Savova,
  et~al.
\newblock 2014.
\newblock Temporal annotation in the clinical domain.
\newblock {\em Transactions of the Association for Computational Linguistics},
  2:143--154.

\bibitem[\protect\citename{Sutton and McCallum}2006]{sutton2006introduction}
Charles Sutton and Andrew McCallum.
\newblock 2006.
\newblock An introduction to conditional random fields for relational learning.
\newblock {\em Introduction to statistical relational learning}, pages 93--128.

\bibitem[\protect\citename{Tai \bgroup et al.\egroup }2015]{tai2015improved}
Kai~Sheng Tai, Richard Socher, and Christopher~D Manning.
\newblock 2015.
\newblock Improved semantic representations from tree-structured long
  short-term memory networks.
\newblock {\em arXiv preprint arXiv:1503.00075}.

\bibitem[\protect\citename{Zhang}2015]{zhang2015deepdive}
Ce~Zhang.
\newblock 2015.
\newblock {\em DeepDive: A Data Management System for Automatic Knowledge Base
  Construction}.
\newblock {Ph.D.} thesis, UW-Madison.

\end{thebibliography}
\bibliographystyle{naaclhlt2016}

\end{document}